\documentclass{article}
\usepackage{amsmath}

\usepackage{PRIMEarxiv}

\usepackage[utf8]{inputenc} 
\usepackage[T1]{fontenc}    
\usepackage{hyperref}       
\usepackage{url}            
\usepackage{booktabs}       
\usepackage{amsfonts}       
\usepackage{nicefrac}       
\usepackage{microtype}      
\usepackage{lipsum}
\usepackage{fancyhdr}       
\usepackage{graphicx}       
\graphicspath{{media/}}     

\title{Few-shot Hate Speech Detection Based on the MindSpore Framework
\thanks{\textit{\underline{Citation}}: 
\textbf{Authors. Title. Pages.... DOI:000000/11111.}} 
}

\author{
  Zhenkai Qin$^{1,2,3}$ \thanks{These authors contributed equally to this work.} \\
  $^{1}$School of Computing and Information \\
  $^{2}$Network Security Research Center \\
  $^{3}$Big Data and Policing Technology Laboratory\\
  Guangxi Police College\\
  Nanning, Guangxi, China \\
  \texttt{qinzhenkai@gxjcxy.edu.cn} \\
  \And
  Dongze Wu \\
  Institute of Software \\
  Chinese Academy of Sciences   \\
  100190,China \\
  \texttt{dongze@isrc.iscas.ac.cn} \\
  \And
  Yuxin Liu \\
  School of Information Technology \\
  Guangxi Police College   \\
  530028,China \\
  \texttt{3105472417@qq.com}  
  \And
  Guifang Yang \\
  School of Information Technology \\
  Guangxi Police College   \\
  530028,China \\
  \texttt{yangguifang@gxjcxy.edu.cn
}
}

\begin{document}
\maketitle

\begin{abstract}
The proliferation of hate speech on social media poses a significant threat to online communities, requiring effective detection systems. While deep learning models have shown promise, their performance often deteriorates in few-shot or low-resource settings due to reliance on large annotated corpora. To address this, we propose \textbf{MS-FSLHate}, a prompt-enhanced neural framework for few-shot hate speech detection implemented on the MindSpore deep learning platform. The model integrates learnable prompt embeddings, a CNN-BiLSTM backbone with attention pooling, and synonym-based adversarial data augmentation to improve generalization. Experimental results on two benchmark datasets—HateXplain and HSOL—demonstrate that our approach outperforms competitive baselines in precision, recall, and F1-score. Additionally, the framework shows high efficiency and scalability, suggesting its suitability for deployment in resource-constrained environments. These findings highlight the potential of combining prompt-based learning with adversarial augmentation for robust and adaptable hate speech detection in few-shot scenarios.
\end{abstract}


\section{Introduction}

In recent years, the proliferation of hate speech on social media platforms has emerged as a critical societal concern. The ease of information dissemination via platforms such as Twitter, Facebook, and Gab has unfortunately enabled the rapid spread of discriminatory, offensive, and hateful content targeting individuals and communities based on race, gender, religion, and other attributes. The prevalence of such content not only threatens online safety and civil discourse but also contributes to offline violence, psychological harm, and the marginalization of vulnerable groups. In response, automatic hate speech detection has become an urgent and highly active area of research within natural language processing (NLP).

Traditional approaches to hate speech detection typically rely on rule-based systems or classical machine learning algorithms such as support vector machines (SVM) and logistic regression. More recently, deep learning-based methods, including convolutional neural networks (CNNs), recurrent neural networks (RNNs), and transformer-based models, have shown substantial promise due to their ability to learn contextual and semantic features from text. Despite their success, these models often assume access to large-scale annotated datasets and may suffer from poor generalization when applied to real-world scenarios with limited or imbalanced data. Moreover, hate speech is highly contextual and subtle, often requiring nuanced understanding and domain-specific cues, which further complicates model development.

One of the most pressing challenges in building reliable hate speech detection systems is the scarcity of high-quality annotated data. Manual labeling is labor-intensive, time-consuming, and ethically sensitive, making large-scale dataset construction difficult, particularly for emerging or under-resourced languages and domains. Additionally, hateful content frequently exhibits long-tail distributions, where truly harmful instances are rare and diverse, exacerbating the class imbalance problem. These constraints highlight the need for methods that are effective in few-shot learning scenarios—where only a small number of labeled examples are available for training.

To address these limitations, we propose a novel framework for few-shot hate speech detection based on prompt-enhanced deep learning, implemented using the MindSpore platform. MindSpore is an open-source deep learning framework developed in China that offers efficient graph-mode execution, strong hardware adaptability, and high modularity for model customization. These properties make it particularly suitable for rapid prototyping and experimentation in research settings with limited computational resources.

Our method introduces a learnable prompt embedding mechanism that prepends task-specific virtual tokens to input sequences, enabling the model to better encode semantic priors even under data-scarce conditions. In addition, we employ adversarial data augmentation via synonym replacement, which increases lexical diversity while preserving semantic consistency. This strategy enhances the model’s robustness to variations in phrasing and vocabulary, which are common in user-generated content. Comprehensive experiments are conducted on two publicly available benchmark datasets—HateXplain and the Hate Speech and Offensive Language Dataset (HSOL)—to demonstrate the effectiveness of the proposed approach.

The main contributions of this paper are as follows:
\begin{itemize}
    \item We propose a novel prompt-based neural architecture for few-shot hate speech detection, built upon the MindSpore framework to leverage its efficient graph-based execution and flexibility.
    \item We introduce an adversarial data augmentation strategy based on synonym replacement to improve model generalization under limited supervision.
    \item We conduct extensive experiments on two benchmark datasets, showing that our method outperforms existing baselines across multiple evaluation metrics in few-shot settings.
\end{itemize}

\section{Related Work}

Automated hate speech detection has received growing attention due to the increasing volume of offensive and abusive content on social media platforms. Early approaches primarily relied on rule-based systems or traditional machine learning classifiers such as support vector machines (SVMs) and logistic regression~\cite{schmidt2017survey}, often supported by handcrafted features or keyword lists. While these methods were computationally efficient, they typically suffered from poor generalization and high false positive rates, particularly when dealing with implicit hate, sarcasm, or code-switching. The advent of deep learning led to substantial improvements in detection performance, with convolutional neural networks (CNNs), recurrent neural networks (RNNs), and more recently, transformer-based models such as BERT and RoBERTa demonstrating strong capabilities in capturing contextual semantics~\cite{zhang2018detecting, macavaney2019hate, plaza2021multi}. Complementary enhancements have included attention-based pooling~\cite{gamback2017using}, multi-task learning~\cite{zampieri2023offensive}, and multimodal architectures that integrate user metadata or visual content~\cite{qureshi2021compromised}. However, the majority of these methods assume access to large-scale annotated datasets and tend to degrade when applied to domain-shifted or data-scarce scenarios.

To alleviate the reliance on large labeled corpora, few-shot learning (FSL) has been actively explored as a data-efficient alternative. Meta-learning approaches such as Prototypical Networks, Matching Networks~\cite{laenen2021episodes}, and MAML~\cite{finn2017model} have shown success in NLP tasks including intent classification and relation extraction~\cite{han2018fewrel}. Nevertheless, applying FSL to hate speech detection remains a largely underexplored direction due to challenges such as class imbalance, semantic ambiguity, and the subtle nature of toxic language. Moreover, conventional fine-tuning of large-scale pre-trained models often leads to overfitting when only a few training samples are available~\cite{yin2019meta}. Recent studies have introduced task-adaptive pretraining, prototype-based regularization, and contrastive learning~\cite{bragg2021flex, gao2020making} to improve few-shot generalization, yet these strategies have not been systematically applied to socially sensitive tasks like hate speech detection.

Parallel to these developments, prompt-based learning has emerged as a powerful paradigm for adapting large language models to downstream tasks with minimal supervision. By reformatting classification problems into cloze-style or masked language modeling templates, prompt tuning leverages the pre-trained knowledge embedded within large models~\cite{liu2023pre}. Early methods focused on manually designed prompts, which were later superseded by learnable alternatives such as soft prompt tuning~\cite{lester2021power}, prefix tuning~\cite{li2021prefix}, and instruction tuning~\cite{wei2021finetuned}. These techniques have achieved competitive performance in various few-shot tasks, including sentiment classification and question answering. However, their application to hate speech detection remains limited. Some recent efforts have considered prompt-based toxicity classification~\cite{wang2022toxicity}, but these studies rarely address challenges related to robustness, fairness, or linguistic diversity, particularly in low-resource settings.

Building on these prior directions, our work proposes a prompt-enhanced neural architecture specifically designed for few-shot hate speech detection. It integrates learnable task-adaptive prompts with a lightweight CNN-BiLSTM backbone and attention pooling to capture both local and long-range dependencies. Furthermore, we incorporate a synonym-based adversarial data augmentation mechanism to improve lexical robustness and semantic generalization. Unlike most prior approaches, our method is implemented on the MindSpore framework, which supports graph-mode execution and efficient deployment on AI-native hardware. This design enables scalable training and practical deployment in resource-constrained environments, positioning our framework as a strong candidate for real-world hate speech moderation under low-data constraints.

\section{Method}

\subsection{Proposed Framework}

The primary goal of this research is to effectively address the challenge of hate speech detection in scenarios with limited labeled data. To this end, we propose a novel few-shot neural architecture named \textbf{MS-FSLHate} (MindSpore-based Few-Shot Learning for Hate Speech), which integrates prompt-enhanced embeddings, convolutional neural networks (CNN), bidirectional Long Short-Term Memory (BiLSTM), and an attention-based pooling strategy. Additionally, the framework incorporates synonym-based adversarial data augmentation and a carefully designed training pipeline tailored for few-shot scenarios. MS-FSLHate is implemented using the MindSpore deep learning platform, leveraging its computational efficiency and deployment scalability.

\begin{figure}[ht]
    \centering
    \includegraphics[width=0.9\linewidth]{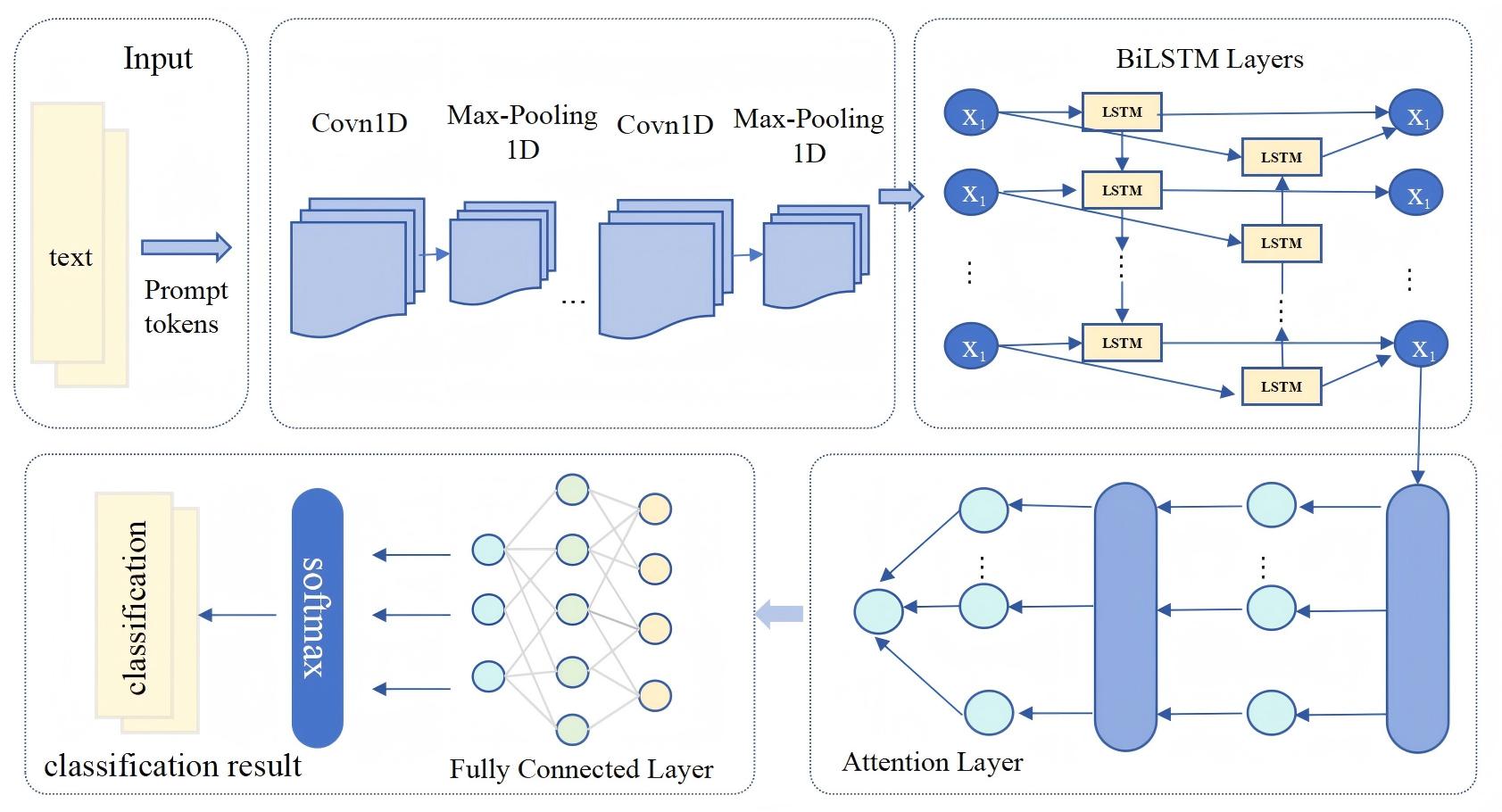}
    \caption{Overview of the proposed MS-FSLHate framework for few-shot hate speech detection.}
    \label{fig:framework}
\end{figure}

Fig.~\ref{fig:framework} presents an overview of the proposed architecture, which comprises four integral components: (1) a prompt-enhanced neural embedding module designed to introduce task-specific semantic guidance, significantly facilitating learning under data scarcity; (2) a convolutional feature extraction module, which captures local contextual features from embedded token sequences; (3) a BiLSTM layer integrated with an attention mechanism, which models sequential dependencies and dynamically emphasizes key semantic elements; and (4) a fully-connected classification layer responsible for predicting the final labels. Furthermore, the training strategy integrates several advanced optimization techniques, including class-weighted loss, synonym replacement for data augmentation, cosine annealing learning rate scheduling, and global gradient clipping to enhance the robustness and stability of the model.

Formally, let us consider an input social media text represented as a sequence of tokens \( X = [x_1, x_2, \dots, x_n] \). Each token is first transformed into a dense vector via an embedding layer. To encode task-aware context, we prepend a set of learnable prompt embeddings to the input sequence. These virtual tokens guide the network to attend to hate-speech-relevant semantics during encoding. The augmented sequence is passed through convolutional layers to extract local features, which are then fed into a BiLSTM module to capture bidirectional dependencies. An attention-based pooling mechanism selectively aggregates important representations, which are subsequently passed through fully-connected layers to yield final predictions across three categories: \textit{normal}, \textit{offensive}, and \textit{hatespeech}.

The effectiveness and generalizability of \textbf{MS-FSLHate} are empirically verified through extensive experiments on benchmark datasets, as described in subsequent sections, showing that it significantly outperforms existing baselines in few-shot hate speech detection scenarios.

\subsection{Prompt-based learning}

Prompt-based learning has emerged as a promising paradigm for addressing few-shot learning challenges in natural language processing tasks. Inspired by recent advances in prompt learning, we introduce a learnable prompt embedding mechanism into our neural architecture to explicitly incorporate task-specific contextual guidance for hate speech detection. The rationale behind prompt embeddings lies in their capability to supply additional semantic cues, significantly improving generalization performance in limited labeled data scenarios.

Formally, given an input token sequence represented as \( X = [x_1, x_2, \dots, x_n] \), each token \( x_i \) is first converted into a dense vector representation via an embedding layer, resulting in the embedding sequence:

\begin{equation}
E = [e_1, e_2, \dots, e_n], \quad e_i \in \mathbb{R}^{d_{emb}}
\end{equation}

where \( d_{emb} \) denotes the embedding dimensionality. To explicitly incorporate task-oriented prior knowledge, we introduce a set of learnable prompt embeddings defined as:

\begin{equation}
P = [p_1, p_2, \dots, p_m], \quad p_j \in \mathbb{R}^{d_{emb}}
\end{equation}

where \( m \) represents the prompt length. The prompt embeddings \( P \) are randomly initialized and optimized jointly with other network parameters during training, allowing the model to dynamically adjust these prompts toward better performance.

To effectively integrate these prompts with the original token embeddings, we concatenate the prompt embeddings \( P \) with the token embeddings \( E \), resulting in an augmented embedding sequence \( E' \):

\begin{equation}
E' = [p_1, p_2, \dots, p_m, e_1, e_2, \dots, e_n]
\end{equation}

This combined embedding sequence encodes explicit task-specific semantics, thereby guiding subsequent neural network modules toward capturing features that are particularly informative for hate speech detection. Specifically, the concatenated sequence is passed through subsequent convolutional and recurrent layers, as described in Sections~\ref{sec:cnn} and~\ref{sec:bilstm_attention}, to extract discriminative features for classification.

By employing this prompt-enhanced embedding strategy, our model effectively leverages contextual prior knowledge to improve feature representation learning, significantly mitigating the inherent difficulties associated with limited labeled training examples in hate speech detection tasks.

\subsection{Convolutional Feature Extractor}
\label{sec:cnn}

To effectively capture local semantic patterns embedded in the augmented input sequence, we employ a convolutional neural network (CNN) as a critical feature extraction component. CNNs have been widely demonstrated to be effective in extracting local semantic representations from textual data, primarily due to their capability of capturing important n-gram features through convolutional kernels.

Formally, given the augmented embedding sequence \( E' = [e'_1, e'_2, \dots, e'_{m+n}] \), as described in Section~3.2, the CNN layer applies convolutional operations over the embedding dimension to produce local semantic features. Specifically, we use a one-dimensional convolutional layer with multiple convolutional kernels of fixed window size. Each convolutional filter slides across the embedded token sequence, computing local feature representations based on contiguous subsets of embeddings. The convolutional operation can be mathematically defined as follows:

\begin{equation}
c_i = f\left(W_{conv} \ast E'_{i:i+k-1} + b_{conv}\right)
\end{equation}

where \(W_{conv}\) and \(b_{conv}\) denote the convolutional kernel weights and biases, respectively; \(\ast\) represents the convolution operation; \(k\) is the convolutional kernel size, and \(f(\cdot)\) denotes a nonlinear activation function, specifically a Rectified Linear Unit (ReLU) in this study.

Subsequently, we apply a max-pooling operation to reduce the dimensionality of feature representations while preserving critical local features. Max-pooling helps to select the most salient local features, enhancing the model's robustness to textual variability. Formally, the max-pooling operation is described as:

\begin{equation}
z_j = \max\{c_{j\times s}, c_{j\times s+1}, \dots, c_{j\times s+s-1}\}
\end{equation}

where \(s\) denotes the pooling size and \(z_j\) represents the pooled feature representation.

Through this convolutional feature extraction procedure, our model effectively captures local semantic dependencies inherent in hate speech contexts, thereby providing informative and discriminative representations as inputs for the subsequent bidirectional recurrent layers.
``

\subsection{BiLSTM with Attention Pooling}
\label{sec:bilstm_attention}

While convolutional layers are effective at capturing local dependencies, they are inherently limited in modeling long-range contextual relationships. To address this limitation, we incorporate a Bidirectional Long Short-Term Memory (BiLSTM) network to capture both forward and backward semantic dependencies across the entire input sequence. BiLSTM networks have proven particularly effective in natural language understanding tasks, especially when sequential and contextual information plays a critical role.

Let the output of the convolutional feature extractor be denoted as a sequence of hidden states:
\begin{equation}
H = [h_1, h_2, \dots, h_T], \quad h_t \in \mathbb{R}^{d_{conv}}
\end{equation}
where \( T \) is the length of the sequence after pooling, and \( d_{conv} \) is the dimension of the convolutional output at each time step.

The BiLSTM processes this sequence in both forward and backward directions, producing a contextualized representation for each position:
\begin{equation}
\overrightarrow{h_t} = \text{LSTM}_{\text{fwd}}(h_t), \quad \overleftarrow{h_t} = \text{LSTM}_{\text{bwd}}(h_t)
\end{equation}
\begin{equation}
\tilde{h_t} = [\overrightarrow{h_t}; \overleftarrow{h_t}] \in \mathbb{R}^{2d_{lstm}}
\end{equation}

Here, \( \tilde{h_t} \) is the concatenated hidden state from both directions at time step \( t \), and \( d_{lstm} \) is the dimensionality of each unidirectional LSTM.

To aggregate the BiLSTM outputs into a fixed-size vector for classification, we apply an attention pooling mechanism that enables the model to weigh different time steps according to their semantic importance to the task. The attention weight \( \alpha_t \) for each hidden state is computed as:

\begin{equation}
\alpha_t = \frac{\exp(w^\top \tilde{h_t})}{\sum_{i=1}^{T} \exp(w^\top \tilde{h_i})}
\end{equation}

where \( w \in \mathbb{R}^{2d_{lstm}} \) is a trainable attention vector. The final context vector \( c \) is obtained as a weighted sum of all hidden states:

\begin{equation}
c = \sum_{t=1}^{T} \alpha_t \tilde{h_t}
\end{equation}

This attention-based aggregation allows the model to focus selectively on informative regions of the sequence that are more indicative of hate speech, while down-weighting less relevant content. Finally, the context vector \( c \) is normalized using a layer normalization operation and passed to a fully connected layer for final classification.

The combined use of BiLSTM and attention pooling provides a powerful mechanism to capture global sequence-level features while preserving the flexibility to highlight critical information, making it particularly well-suited for hate speech detection in noisy and short text inputs.

\subsection{MindSpore Implementation and Adversarial Augmentation}
\label{sec:mindspore-ada}

To ensure training efficiency and deployment compatibility, the entire model is implemented using the MindSpore deep learning framework. MindSpore provides a computational graph-based execution mode, efficient hardware utilization, and seamless integration with model training and evaluation pipelines. Its high modularity and native support for dynamic networks make it a suitable choice for research on natural language processing tasks, particularly in low-resource or custom model scenarios.

In our implementation, all neural components---including the prompt embedding layer, convolutional feature extractor, BiLSTM module, attention pooling, and classification head---are constructed using MindSpore's high-level neural cell APIs. The model is trained in \texttt{GRAPH\_MODE} on CPU devices, demonstrating the framework's versatility across hardware platforms. The optimizer and loss function are integrated into a custom training step using the \texttt{TrainOneStepCell}, which also includes gradient clipping for training stability. Cosine annealing is employed for learning rate scheduling, and class imbalance is mitigated through weighted cross-entropy loss, although these training strategies are discussed in detail in Section 5.2.

To enhance the robustness and generalization of the model in few-shot settings, we apply adversarial data augmentation (ADA) in the form of synonym replacement. Specifically, for each input token sequence, we randomly replace certain tokens with their synonyms retrieved from WordNet, based on a predefined replacement probability. This augmentation strategy increases lexical diversity in the training data while preserving semantic consistency, thus simulating potential adversarial shifts in user-generated content.

Let \( X = [x_1, x_2, \dots, x_n] \) be the original input token sequence. The ADA-augmented sequence \( \hat{X} = [\hat{x}_1, \hat{x}_2, \dots, \hat{x}_n] \) is generated such that for each token \( x_i \), with probability \( p \), it is replaced by a randomly sampled synonym \( \hat{x}_i \in \text{Synonyms}(x_i) \). Otherwise, \( \hat{x}_i = x_i \). This stochastic process can be expressed as:

\begin{equation}
\hat{x}_i = 
\begin{cases}
\text{RandomSynonym}(x_i), & \text{with probability } p \\
x_i, & \text{otherwise}
\end{cases}
\end{equation}

By combining MindSpore's computational efficiency with ADA's augmentation strength, our model is better equipped to handle noise, variation, and underrepresented linguistic patterns in the hate speech detection task.

\section{Empirical Research}
\subsection{Datasets}

To evaluate the effectiveness and generalizability of the proposed model, we conduct experiments on two publicly available benchmark datasets widely used for hate speech detection: HateXplain and the Hate Speech and Offensive Language Dataset (HSOL). Both datasets follow a three-class labeling scheme, which categorizes text into \textit{hate speech}, \textit{offensive language}, and \textit{neutral} or \textit{normal}, thereby aligning with the task formulation of this study. These datasets differ in their source domains, annotation strategies, and linguistic characteristics, offering a complementary basis for evaluating model performance under diverse conditions.

HateXplain is a large-scale dataset jointly developed by the Indian Institute of Technology Kharagpur and the University of Hamburg. It contains 20,148 posts collected from Twitter and Gab, annotated via Amazon Mechanical Turk. Each post is labeled along three dimensions: (1) a primary classification label (\textit{hate speech}, \textit{offensive language}, or \textit{normal}); (2) the targeted community referenced in the content; and (3) an annotator-provided rationale indicating the justification for the assigned label. In this work, we exclusively utilize the primary classification labels to maintain consistency with standard hate speech detection tasks. The dataset’s diverse source platforms and rich annotation schema make it a robust benchmark for evaluating not only detection accuracy but also the fairness and interpretability of machine learning models.

The Hate Speech and Offensive Language Dataset (HSOL), introduced by Davidson et al., comprises 24,802 English-language tweets that were manually labeled as \textit{hate speech}, \textit{offensive language}, or \textit{neither}. The tweets were collected using a predefined lexicon of hate-related terms and annotated by trained human coders. Although the dataset contains language that may be considered racist, sexist, or otherwise derogatory, it is representative of the toxic language frequently observed on social media platforms. In our experiments, we adopt the original label definitions to retain the distinction between hateful and merely offensive content.

Table~\ref{tab:dataset-summary} summarizes the key characteristics of both datasets.

\begin{table}[htbp]
    \centering
    \caption{Summary of the datasets used in our experiments.}
    \label{tab:dataset-summary}
    \begin{tabular}{lcccc}
        \toprule
        \textbf{Dataset} & \textbf{Source} & \textbf{Samples} & \textbf{Label Types} & \textbf{Language} \\
        \midrule
        HateXplain & Twitter + Gab & 20,148 & Hate / Offensive / Normal & English \\
        HSOL       & Twitter        & 24,802 & Hate / Offensive / Neither & English \\
        \bottomrule
    \end{tabular}
\end{table}

Both datasets pose inherent challenges such as class imbalance, informal and noisy language, and subtle semantic differences between offensive and hateful expressions. HateXplain, with its multi-perspective annotations, is especially suitable for evaluating interpretability and fairness. In contrast, HSOL offers a large-scale, keyword-triggered corpus that is ideal for assessing robustness under real-world distributional noise. Together, these datasets provide a comprehensive foundation for benchmarking hate speech detection models in both few-shot and domain-generalization scenarios.

\subsection{Experimental Setup}

All experiments are implemented using the MindSpore deep learning framework (version~\texttt{2.5.0}), running in \texttt{GRAPH\_MODE} on a CPU-based environment. MindSpore provides efficient computation graphs and flexible neural cell APIs, making it well-suited for tasks that require architectural customization and scalability, such as few-shot hate speech detection.

The model architecture and training procedures follow the design described in Section~\ref{sec:methodology}. The entire dataset is pre-tokenized, and a vocabulary of 15,000 tokens is built from the combined training and validation sets. Each input sequence is either truncated or padded to a fixed length of 128 tokens. We apply synonym-based adversarial data augmentation to the training data using WordNet, with a token-level replacement probability of 10\%.

\begin{table}[htbp]
    \centering
    \caption{Summary of hyperparameter settings.}
    \label{tab:exp-settings}
    \begin{tabular}{ll}
        \toprule
        \textbf{Parameter} & \textbf{Value} \\
        \midrule
        Embedding dimension & 300 \\
        Hidden dimension (LSTM) & 256 \\
        Prompt length & 10 \\
        Sequence length & 128 \\
        Batch size & 32 \\
        Number of epochs & 3 \\
        Optimizer & AdamW \\
        Learning rate (initial) & 0.0005 \\
        Learning rate (minimum) & $1 \times 10^{-5}$ \\
        Weight decay & $1 \times 10^{-5}$ \\
        Learning rate schedule & Cosine annealing \\
        Dropout keep probability & 0.7 \\
        Dropout rate (LSTM) & 0.2 \\
        Gradient clipping norm & 1.0 \\
        Data augmentation & Synonym replacement (probability 0.1) \\
        \bottomrule
    \end{tabular}
\end{table}

To simulate a low-resource setting, the training and validation subsets are merged during training. The training process incorporates several strategies to enhance stability and generalization, including cosine annealing learning rate scheduling and global gradient clipping. Evaluation is conducted using per-class precision, recall, F1-score. To ensure result consistency, each experiment is repeated across three random seeds, and the averaged results are reported.

Table~\ref{tab:exp-settings} summarizes the key hyperparameters used in our experiments.

\subsection{Results and Analysis}

To evaluate the effectiveness of the proposed model, we compare the performance of \textbf{MS-FSLHate} with several widely used baselines for text classification: a standard Recurrent Neural Network (RNN), Hierarchical Attention Network (HAN), and a bidirectional GRU with attention mechanism (BiGRU-ATT). The evaluation is conducted on two benchmark datasets: HateXplain and HSOL. For each model, we report precision, recall, and F1-score, averaged across three runs to ensure robustness. The results are summarized in Table~\ref{tab:results_combined}.

\begin{table}[htbp]
\centering
\caption{Performance comparison on HateXplain and HSOL datasets.}
\renewcommand{\arraystretch}{1.5}
\label{tab:results_combined}
\begin{tabular}{lcccccc}
\toprule
 & \multicolumn{3}{c}{\textbf{HateXplain Dataset}} & \multicolumn{3}{c}{\textbf{HSOL Dataset}} \\
\cmidrule(lr){2-4} \cmidrule(lr){5-7}
\textbf{Model} & \textbf{P(\%)} & \textbf{R(\%)} & \textbf{F1(\%)} & \textbf{P(\%)} & \textbf{R(\%)} & \textbf{F1(\%)} \\
\midrule
RNN           & 39.04  & 40.16  & 39.33  & 69.71  & 75.39  & 69.33 \\
HAN           & 62.34  & 60.05  & 60.61  & 62.19  & 76.44  & 64.91 \\
BiGRU-ATT     & 60.78  & 60.52  & 60.55  & 65.44  & 70.61  & 66.90 \\
\textbf{MS-FSLHate (Ours)} & \textbf{65.45} & \textbf{65.78} & \textbf{65.56} & \textbf{85.16} & \textbf{85.21} & \textbf{85.16} \\
\bottomrule
\end{tabular}
\end{table}

As shown in Table~\ref{tab:results_combined}, \textbf{MS-FSLHate} significantly outperforms all baseline models on both datasets across all evaluation metrics. On the HateXplain dataset, it achieves an F1-score of 65.56\%, exceeding the next best model (HAN) by approximately 5 percentage points. The performance gains are consistent across precision and recall, indicating balanced improvement rather than bias toward a specific metric.

On the HSOL dataset, the improvements are even more pronounced. \textbf{MS-FSLHate} achieves an F1-score of 85.16\%, outperforming BiGRU-ATT by over 18 points and HAN by more than 20 points. This highlights the model’s strong generalization capability in noisy, real-world social media data, particularly in detecting nuanced and diverse expressions of offensive or hateful language.

\begin{figure}[htbp]
    \centering
    \includegraphics[width=\linewidth]{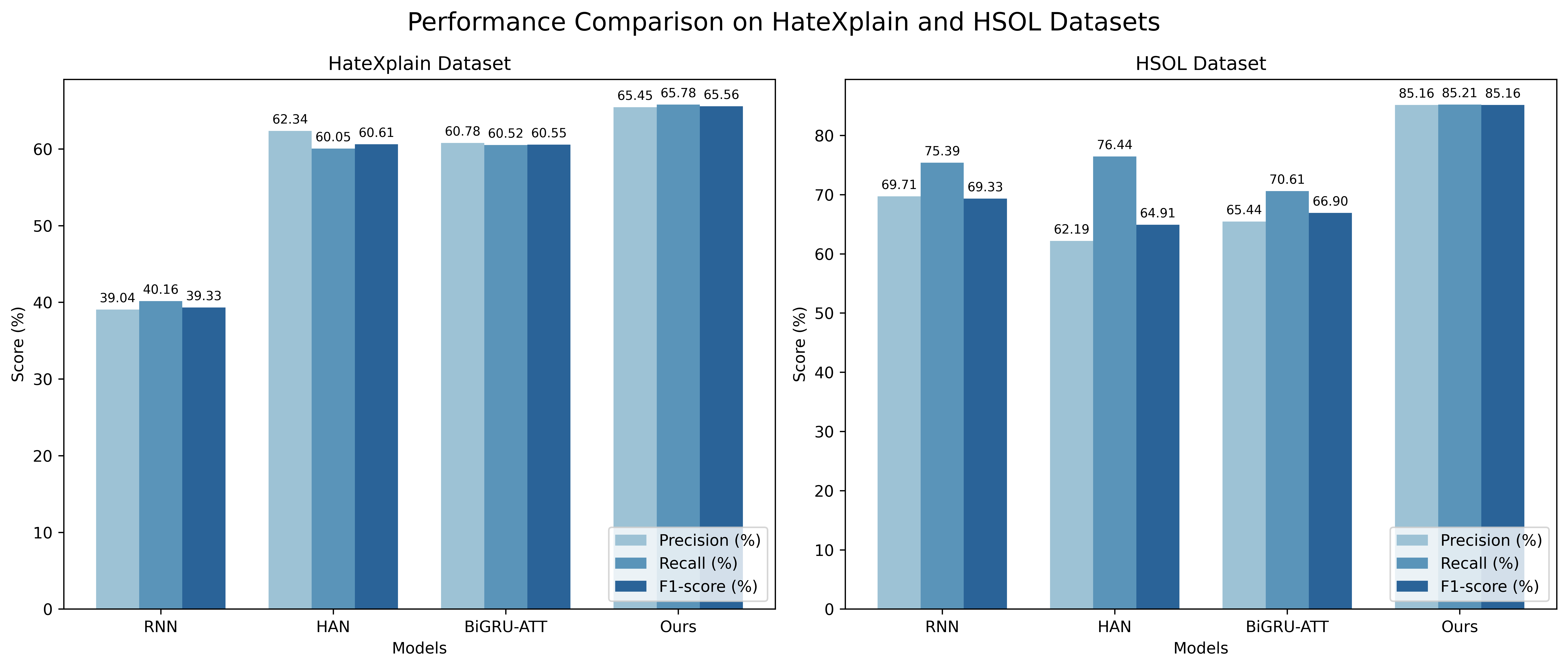}
    \caption{Performance comparison across models on HateXplain (left) and HSOL (right) datasets. Metrics include precision, recall, and F1-score. \textbf{MS-FSLHate} demonstrates consistent superiority across all metrics.}
    \label{fig:performance_bar}
\end{figure}

In addition to the numerical results, Figure~\ref{fig:performance_bar} provides a visual comparison of model performance across both datasets. The bar charts clearly illustrate the consistent advantage of \textbf{MS-FSLHate} across all metrics. The performance gap is especially visible in the HSOL dataset, where \textbf{MS-FSLHate} maintains high precision and recall simultaneously—demonstrating its ability to balance accurate classification with low false negative rates. The visualizations also underscore the performance limitations of traditional models such as RNN, especially when applied to semantically complex data like HateXplain.

The superior performance of \textbf{MS-FSLHate} can be attributed to the synergy of its core design components: prompt-enhanced embeddings provide task-specific guidance, convolutional layers extract robust local patterns, and the attention-pooled BiLSTM effectively captures long-range semantic dependencies. Additionally, the use of synonym-based adversarial augmentation enhances generalization, particularly in detecting underrepresented or context-sensitive forms of hate speech.

These results collectively demonstrate that \textbf{MS-FSLHate} is not only effective in few-shot settings but also adaptable across domains with varied data distributions and linguistic complexity.

\subsection{Ablation Studies}

To evaluate the contribution of individual components in the proposed architecture, we conducted a series of ablation experiments on the HateXplain dataset. Specifically, we assessed the performance impact of removing the key modules within \textbf{MS-FSLHate}, including the prompt embedding, adversarial data augmentation, and attention pooling components. The model's effectiveness was measured using precision, recall, and F1-score. Table~\ref{tab:ablation_study} summarizes the results obtained from each ablated configuration.

\begin{table}[htbp]
\centering
\caption{Ablation study results on the HateXplain dataset. All metrics are reported in percentages.}
\label{tab:ablation_study}
\begin{tabular}{lccc}
\toprule
\textbf{Model Variant} & \textbf{Precision (\%)} & \textbf{Recall (\%)} & \textbf{F1-score (\%)} \\
\midrule
Full model (MS-FSLHate)     & \textbf{65.45} & \textbf{65.78} & \textbf{65.56} \\
w/o Prompt                  & 61.06          & 59.64          & 60.16          \\
w/o Augmentation            & 62.22          & 63.75          & 62.63          \\
w/o Attention               & 64.06          & 59.17          & 59.67          \\
w/o Prompt \& Augmentation  & 65.44          & 56.82          & 57.74          \\
\bottomrule
\end{tabular}
\end{table}

As shown in Table~\ref{tab:ablation_study}, removing the adversarial data augmentation module results in a 3.2 percentage point decrease in precision and a 2.9 point reduction in F1-score. This highlights the importance of training data diversity in improving the model’s robustness. Excluding the attention mechanism leads to a substantial drop in recall (6.6 points) and a 5.9-point reduction in F1-score, suggesting that attention pooling contributes meaningfully to capturing long-range semantic dependencies, though its impact is relatively moderate.

\begin{figure}[ht]
    \centering
    \includegraphics[width=0.9\linewidth]{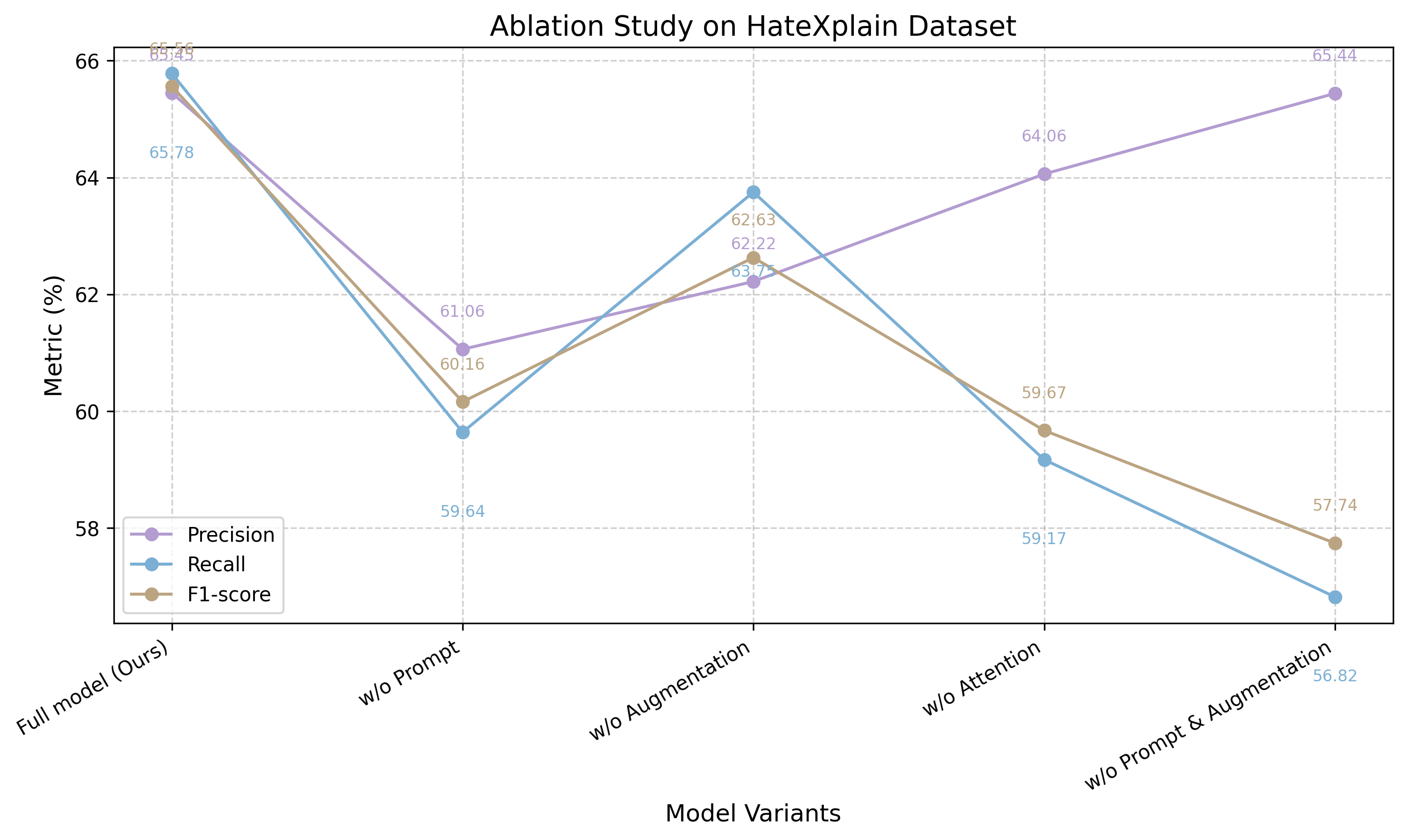}
    \caption{Ablation study on the HateXplain dataset. The line chart illustrates variations in precision, recall, and F1-score across different model variants of MS-FSLHate.}
    \label{fig:ablation_curve}
\end{figure}

The most significant degradation occurs when both prompt embedding and augmentation are removed, with recall dropping by 8.96 points and F1-score decreasing by 7.82 points. This result strongly supports the hypothesis that these components offer complementary benefits and are jointly responsible for enhancing generalization in few-shot learning settings.

Figure~\ref{fig:ablation_curve} provides a visual representation of these trends. The full \textbf{MS-FSLHate} model consistently yields the highest performance across all metrics. Precision remains relatively stable across variants, while recall appears more sensitive to the removal of prompts and data augmentation. This visual pattern reinforces the quantitative findings, confirming that each architectural module contributes distinctly to the overall model effectiveness.

In conclusion, the ablation results empirically demonstrate that prompt embedding and synonym-based augmentation are critical for achieving high performance in few-shot hate speech detection. While the attention mechanism also contributes positively, its relative impact is less pronounced. These findings highlight the architectural efficiency and design rationality of \textbf{MS-FSLHate}, particularly under data-scarce conditions.

\section{Conclusion}

This study proposed \textbf{MS-FSLHate}, a novel prompt-based framework for few-shot hate speech detection, implemented using the MindSpore deep learning platform~\cite{tong2021study}. The architecture comprises three core components: a learnable prompt embedding module that facilitates task-aware semantic representation, a CNN-BiLSTM backbone with attention pooling for capturing both local and global contextual features, and a synonym-based adversarial data augmentation strategy that enhances generalization in low-resource settings~\cite{gupta2024harmful}.

Extensive experiments conducted on two publicly available benchmark datasets—HateXplain~\cite{bhattacharyabattling} and the Hate Speech and Offensive Language Dataset (HSOL)~\cite{alsekait2024semantic}—demonstrate that \textbf{MS-FSLHate} consistently outperforms competitive baselines in terms of precision, recall, and F1-score. Ablation studies further confirm the effectiveness of prompt embeddings and adversarial augmentation in enhancing model robustness under data-scarce conditions. These results validate the framework's capacity to address key challenges in hate speech detection, including data scarcity, class imbalance, and semantic ambiguity.

Furthermore, the integration with the MindSpore framework—featuring efficient graph-mode execution and modular architecture—enables scalable training and streamlined deployment~\cite{huawei2022deep}. This ensures that \textbf{MS-FSLHate} remains both performant and resource-efficient, making it viable for deployment in real-world environments with limited computational resources or edge AI requirements.

In future work, we plan to extend \textbf{MS-FSLHate} to multilingual and code-mixed hate speech detection tasks, where linguistic diversity and domain shift pose additional challenges. We also aim to incorporate model interpretability techniques, such as rationale extraction or attention visualization, to enhance transparency and facilitate ethical deployment. In addition, the integration of advanced prompt-tuning and in-context learning strategies will be investigated to further boost performance in extremely low-resource scenarios~\cite{wahidur2024enhancing,meshkin2024harnessing}.

\section*{Acknowledgments}
Thanks for the support provided by the MindSpore Community.

\bibliographystyle{unsrt}  
\bibliography{references}

\end{document}